\documentclass[final,5p,times,twocolumn]{elsarticle}
\usepackage{amssymb}
\usepackage{amsmath}
\usepackage{amsfonts}
\usepackage{picins}

\usepackage{graphicx}
\usepackage{caption}
\usepackage{subcaption}
\graphicspath{{image/}}
\DeclareGraphicsExtensions{.pdf,.jpeg,.png,eps}
\usepackage{multirow}
\usepackage{color}

\usepackage{lineno}

\begin{document}

\begin{frontmatter}



\title{Adaptive Deep Metric Embeddings for Person Re-Identification under Occlusions}
\author[a]{Wanxiang Yang}

\author[a]{Yan Yan\corref{mycorrespondingauthor}}
%\ead{Tel./fax: +86 5922580063.}
%\ead{yanyan@xmu.edu.cn}
\author[b]{Si Chen}
%\author[a]{Hanzi Wang}

\address[a]{Fujian Key Laboratory of Sensing and Computing for Smart City, School of Information Science and Engineering,
Xiamen University, \\Xiamen 361005, Fujian, China}
\address[b]{School of Computer and Information Engineering,
Xiamen University of Technology, \\Xiamen 361024, Fujian, China }

\begin{abstract}
	Person re-identification (ReID) under occlusions is a challenging problem in video surveillance. Most of existing person ReID methods take advantage of local features to deal with occlusions. However, these methods usually independently extract features from the local regions of an image without considering the relationship among different local regions. In this paper, we propose a novel person ReID method, which learns the spatial dependencies between the local regions and extracts the discriminative feature representation of the pedestrian image based on Long Short-Term Memory (LSTM), dealing with the problem of  occlusions. In particular, we propose a novel loss (termed the adaptive nearest neighbor loss) based on the classification uncertainty to effectively reduce intra-class variations while enlarging inter-class differences within the adaptive neighborhood of the sample. The proposed loss enables the deep neural network to adaptively learn discriminative metric embeddings, which significantly improve the generalization capability of recognizing unseen person identities. Extensive comparative evaluations on challenging person ReID datasets demonstrate the significantly improved performance of the proposed method compared with several state-of-the-art methods.
\end{abstract}
\end{frontmatter}
\section{Introduction}
	Matching pedestrians across different camera views, known as person re-identification (ReID), is a challenging task in computer vision \cite{ Lin2016Person, Huang2017DeepDiff, Fang2017Perceptual, Dong2018Person, Cheng2017Discriminative, Zheng2013Reidentification}. One key challenge of person ReID is the significant appearance variations caused by occlusions in pedestrian images.

  There are two major components in the conventional person ReID methods: 1) the effective feature descriptor (such as SCNCD \cite{Yang2014Salient}, gBiCov \cite{Ma2014Covariance}, and LOMO \cite{Liao2015Person}) to characterize the pedestrian image, and 2) the suitable metric (such as LADF \cite{li2013learning}, KISSME \cite{koestinger2012large}, and XQDA \cite{Liao2015Person}) to compare the similarity between pedestrian images.
	With the popularity of deep learning, several deep learning based person ReID methods \cite{Karanam2016A, Ding2015Deep, chen2017beyond, Ustinova2017Multi, Liu2017End} that effectively incorporate these two components into an integrated framework have been developed. Generally, the deep learning based methods automatically learn discriminative image representations based on large scale image data, which have shown to be highly robust to pedestrian appearance variations.

	Partially occluded pedestrians are ubiquitous in person ReID. To deal with the problem of occlusions, several methods \cite{Yi2014Deep}, \cite{Cheng2016Person} have been proposed by taking advantage of the part-based network architectures to learn representations from different local regions of pedestrian images. For example, in \cite{Yi2014Deep}, the authors firstly split the pedestrian images into three overlapping local regions, and then apply a three-channel Convolutional Neural Network (CNN) architecture to learn discriminative local features from these regions. However, this method may suffer from the problem of spatial misalignment (recall that the local features are separately learned). Recently, Zhong et al. \cite{Zhong2017Random} propose to perform data augmentation with random erasing, which addresses the problem of occlusions to some extent. However, this method does not exploit the relationship of spatial structure in the pedestrian image, which can be beneficial for person ReID.

	Recently, Recurrent Neural Network (RNN) has shown the powerfulness in handling sequential data due to its great capability of storing the representations of recent inputs. As an effective RNN, Long Short-Term Memory (LSTM) \cite{palangi2016deep} can properly capture the temporal/spatial dependencies.
	In this paper, inspired by the success of LSTM, we propose to model the spatial dependencies among different local regions of  pedestrian images based on LSTM to handle the problem of occlusions. By making use of the internal gating mechanism of the LSTM cells, the proposed method effectively extracts the intrinsic feature representation by memorizing the spatial correlations and ignoring confusing distractors (i.e., occluded local regions),
   %the proposed method can effectively memorize the spatial correlations and automatically encode the spatial information,
   thus leading to performance improvements for person ReID under occlusions.

	The loss function plays an important role in deep learning for the task of person ReID, which aims to learn separable and discriminative deep features. A commonly used loss function, termed the triplet loss \cite{Schroff2015FaceNet}, can significantly improve the capability of distingushing different classes. However, how to generate high-quality triplets (i.e., hard triplet mining) to ensure training efficiency is not a trivial task, since many triplets are uninformative. Moreover, the triplet loss often suffers from the problems of slow convergence and poor local optima, partially due to the fact that a single triplet only considers the pairwise distance between the anchor and the positive (negative) sample.

	In this paper, to overcome the above problems, we propose an Adaptive Nearest Neighbor (ANN) loss based on the classification uncertainty, maintaining a large margin between the inter-class distance and the intra-class distance within the adaptive neighborhood of each sample. In fact, the ANN loss can be viewed as the generalization of the triplet loss. Compared with the triplet loss, the proposed loss effectively exploits the neighborhood information (the distance between the anchor and the neighborhood of each positive/negative sample) of the training data. Therefore, the selected triplets are informative, making the training process quickly converge.
	
	The main contributions are summarized as follows. 1) We propose to exploit the spatial dependencies between the local regions of  pedestrian images based on LSTM, which significantly improves the performance of person ReID under occlusions. LSTM effectively memorizes the spatial correlations and automatically encodes the spatial information so as to reduce the noises caused by occlusions. 2) We develop an adaptive nearest neighbor (ANN) loss, which takes advantage of the neighborhood information to learn adaptive deep metric embeddings based on the classification uncertainty. Experimental results show the superiority of the proposed method compared with the state-of-the-art methods on several challenging person ReID datasets.
	
\section{The proposed method}
	In this section, we firstly introduce the overall framework of the proposed method in Section 2.1. Then, the spatial encoded local features which exploit the spatial dependencies based on LSTM, are given in Section 2.2. Finally, the proposed ANN loss is formulated in Section 2.3.
\begin{figure}[t]
  \centerline{\includegraphics[width=8.5cm]{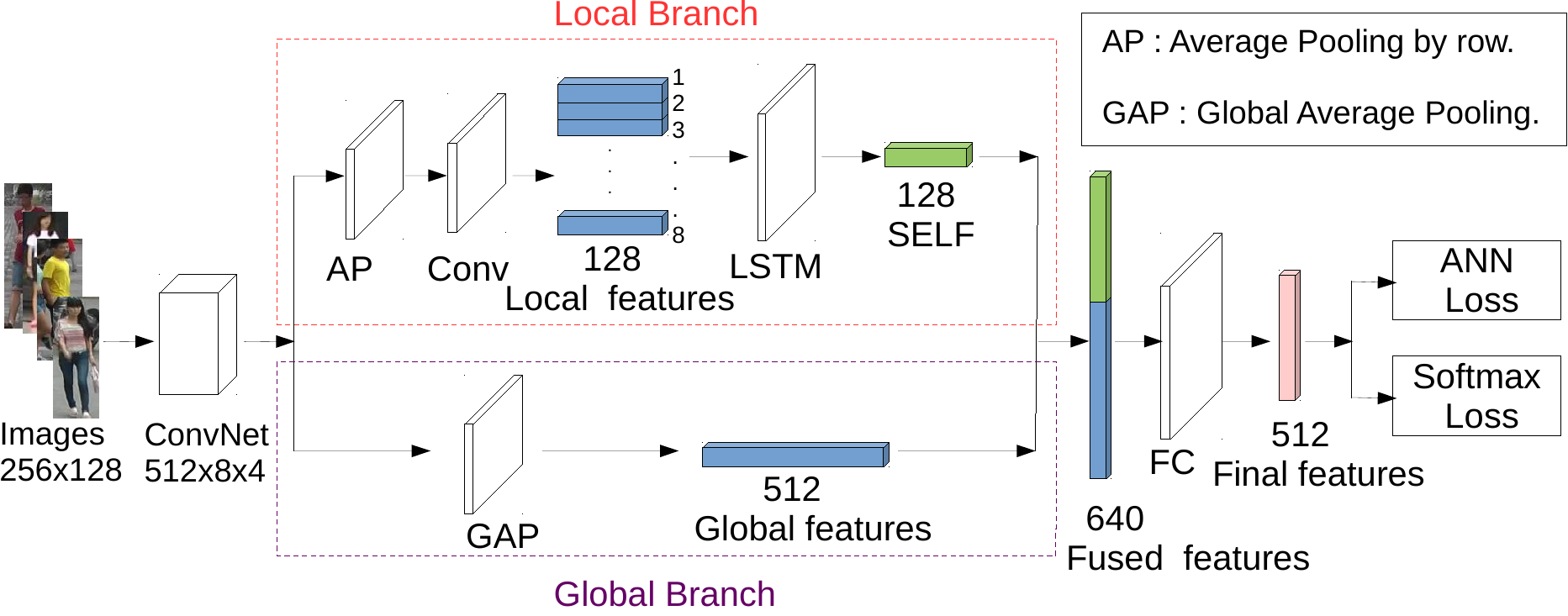}}
  \captionsetup{justification=centering}
  \caption{The overall framework of the proposed method.}\label{fig:arch}
\end{figure}
\subsection{Overall framework}
	The overall framework of the proposed method is illustrated in Fig.~\ref{fig:arch}. For each input pedestrian image $I$, we firstly use a base convolutional neural network (in this paper, ResNet \cite{He2016Deep} is used due to its superiority), to extract mid-level convolutional feature maps. For notational simplicity, we refer to the output of the last convolutional layer of ResNet as $g(I)$ for each input image. Specifically, the last convolutional layer is denoted as $g \in \mathbb{R}^{C \times H \times W }$ ($C=512$, $H=8$, $W=4$ in this paper), where $H$ and $W$ denote the spatial size (i.e., height and width) of the last convolutional layer and $C$ is the number of feature channels. Secondly, the global features and spatial encoded local features (SELF) are respectively extracted in the global and local branches. In the global branch, we extract the global features by applying global average pooling (GAP) to the mid-level feature maps to capture high-level semantics. Meanwhile, in the local branch, the local features are extracted by applying average pooling (AP) to each row of the mid-level feature maps, and then using a $1\times1$ convolutional layer to reduce the number of feature channels from $C$ to $c$ ($c=128$ in this paper). These local features are fed into the LSTM layer to learn the spatial dependencies between different local regions of the input pedestrian image (see Section 2.2). The output of LSTM layer is SELF, which is represented as $L \in \mathbb{R} ^{c}$. Thirdly, the global features and SELF are concatenated to represent the pedestrian image, which contains complementary information from the different levels of semantics, following a fully-connected (FC) layer to get a compact representation. Finally, the deep neural network is jointly optimized by the softmax loss and the proposed adaptive nearest neighbor loss (see Section 2.3).
\subsection{Spatial encoded local features (SELF)}
%	To obtain effective spatial encoded local features (SELF), a large number of partially occluded images are required for training the deep neural network. However, the number of partially occluded images in the existing person ReID datasets is usually limited. Therefore, in this paper, we increase the number of training images by firstly randomly selecting a rectangle region and then reassigning the pixels within the region with random values. In this way, partially occluded images with different levels of occlusion can be generated to simulate the occluded pedestrain images in natural scenes.

	In this section, SELF is extracted to effectively capture the dependencies of spatial structure in the pedestrian image. Specifically, the pedestrian image is decomposed into a sequence of local regions from head to foot, where each local region has a relatively fixed position due to the prior knowledge about the human body structure. Based on the local regions, different local features are extracted accordingly. In this manner, all local features are treated as a spatial sequence. Following that, the sequence data consisting of all the local features are represented as $S_{t}\in \mathbb{R} ^{c}$ ($t =1, \cdots, H$), where $S_{t}$ is the local features in each row and $H$ denotes the sequence length (i.e., the number of local regions).
    The LSTM layer sequentially accepts the input local features and the hidden state $h_{t} \in \mathbb{R} ^{e}$ at each step $t$ is obtained using the following equations ($e$ is the number of hidden units of the LSTM layer. In this paper, we empirically set $e$ to be equal to $c$).
\begin{equation}\label{eq:gate}
\left(
  \begin{array}{c}
    i_{t} \\
    f_{t} \\
    o_{t} \\
    g_{t} \\
  \end{array}
\right)
=
\left(
  \begin{array}{ccc}
    sigm\\
    sigm\\
    sigm \\
    tanh \\
  \end{array}
\right)
W_{L}
\left(
  \begin{array}{ccc}
    S_{t}\\
    h_{t-1}\\
  \end{array}
\right),
\end{equation}

\begin{equation}\label{eq:ct}
d_{t} = f_{t} \odot d_{t-1}+i_{t} \odot g_{t},
\end{equation}
\begin{equation}\label{eq:hidden}
h_{t} = o_{t} \odot tanh(d_{t}),
\end{equation}
where $i_{t} \in \mathbb{R}^{e}$, $f_{t} \in \mathbb{R}^{e}$, $o_{t} \in \mathbb{R}^{e}$, $g_{t} \in \mathbb{R}^{e}$ and $d_{t} \in \mathbb{R}^{e}$ are the input gate, forget gate, output gate, cell state candidate and cell state, respectively. $sigm \in \mathbb{R}^{e}$ and $tanh \in \mathbb{R}^{e}$ denote the non-linear activation functions (i.e., the sigmoid function and the tanh function), which are applied in element-wise. $W_{L} \in \mathbb{R}^{4e \times (c+e)}$ denotes the weight matrix of the LSTM layer. $\odot$ denotes the element-wise multiplication.
	 	
	 From Eq. (\ref{eq:gate}), $i_{t} $, $f_{t}$ and $o_{t}$ decide which information will be updated, thrown away and outputted, respectively, according to the previous hidden state $h_{t-1}$ and the current input $S_{t}$. A $tanh$ activation layer creates the cell candidate state $g_{t}$. In Eq. (\ref{eq:ct}), the LSTM layer updates the old cell state $d_{t-1}$ by firstly multiplying $f_{t}$ (throw away the old information), and adding the cell candidate state scaled by $i_{t}$ (update the information), to obtain the cell state $d_{t}$.
	In Eq. (\ref{eq:hidden}), the LSTM layer passes the cell state $d_{t}$ through $tanh$ (constrain the output values between $-1$ and $1$) and multiplies it by the output gate $o_{t}$ (output the critical information and reduce the noises), to obtain the hidden state $h_{t}$.

	The final hidden state of the LSTM layer is SELF (i.e., $L=h_{H}$, where $L \in \mathbb{R} ^{c}$), which effectively captures the spatial relationship between different local regions. Since partial occlusions only affect some local regions, we exploit the intrinsic relationship between different local regions (recall that the occluded regions are considered as noises that can be filtered by LSTM) to alleviate the problem of occlusions. Therefore, the SELF extracted by the LSTM layer is highly robust to occlusions.
\subsection{Adaptive nearest neighbor loss}
As we discuss previously, hard triplet mining is critical for the triplet loss (note that most triplets are uninformative).
%In addition, the triplet only considers the local information (relative distances are used in the triplet), leading to the problems of slow convergence and poor local optima.
Although some methods \cite{Hermans2017In,Song2016Deep} have been developed for hard triplet mining, the selected triplets only exploit pairwise distance information, which may result in the local optima.

Motivated by the above issues, we propose a novel loss function, termed the adaptive nearest neighbor (ANN) loss, which effectively takes advantage of the neighborhood information to enlarge the inter-class dispersion while preserving the intra-class compactness. To be specific, the ANN loss is defined as follows,	
%Motivated by the aforementioned issues, we propose the ANN that incorporates the moderate triplets mining criterion and utilizes the difficult triplets as many as possible, as follows.		
%	Traditional triplet loss suffers from the problem that the  number of triplets cubically increases with the number of samples. Therefore, hard triplet mining is critical for the triplet loss, since most triplets are trivial and uninformative.
%	In \cite{Hermans2017In}, Hermans et al. introduce a variant of triplet loss termed batch hard loss that seletcts the hardest positive and the hardest negative samples within the batch to constitute the triplets, which can be considered as moderate triplets. However, due to the batch size is limited, the selected triplets only exploit partial distance information, which may result in the local optimal. In \cite{Song2016Deep}, Song et al. propose a lifted structured loss based on all positive and negative pairs of samples in the batch, which takes full advantage of the samples. However, the batch still contains many trivial and uninformative triplets.
\begin{equation}\label{eq:Ln}
L_{ANN}=\sum_{a=1}^{B}[ m+ D_{ap}  - D_{an} ]_{+},
\end{equation}
\begin{equation}\label{eq:Dap}
D_{ap} = \frac{1}{K_a}\sum_{k=1}^{K_a}\left \|  f(I_{a})-f(I_{pk})   \right \| _{2}^{2},
\end{equation}
\begin{equation}\label{eq:Dan}
D_{an} = \frac{1}{K_a}\sum_{k=1}^{K_a}\left \|  f(I_{a})-f(I_{nk})   \right \| _{2}^{2},
\end{equation}
where $[\cdot]_{+}$ denotes the hinge loss. $B$ is the number of training samples. $f(\cdot )$ is the function that maps the raw image to the metric embedding representation. $I_a$, $I_{pk}$ and $I_{nk}$ represent  the anchor sample,  positive sample, and negative sample, respectively.
 $D_{ap}$ and $D_{an}$ respectively denote the average distance between the anchor sample $I_a$ and the $K_a$ hardest positive samples (i.e, the $K_a$ farthest positive samples), and the average distance between $I_a$ and the $K_a$ hardest negative samples (i.e., the $K_a$ closest negative samples). $m$ is a margin to keep the separation between positive and negative pairs. $\left \| \cdot \right \|_{2}$ denotes the Euclidean distance. Note that $K_a$ denotes the number of positive/negative samples in the neighborhood of the anchor sample $I_a$.

In this paper, instead of fixing $K_a$ to be a constant value, we adaptively set $K_a$ based on the classification uncertainty. That is, $K_a$ is formulated as follows:
\begin{equation}\label{eq:sizeK}
K_a = \max(\left \lfloor H_{a} \right \rfloor,K_{0}),
\end{equation}
where $H_{a}=-\sum_{j=1}^{N}p_{a}^{j}log(p_{a}^{j})$ denotes the classification uncertainty of the anchor sample $I_a$. Here, $p_{a}^{j}$ is the probability that the sample $I_a$ belongs to the $j$-th class according to a softmax layer and $N$ is the number of classes.$\left \lfloor \cdot \right \rfloor$ denotes the ceil operation. $K_{0}$ is a constant, denoting the minimum number of nearest neighbors (we set $K_{0}$ to be 1 in this paper).

    The classification uncertainty $H_{a}$ measures the confidence of classification based on the softmax classifier, which intrinsically characterizes the global data distribution. When the value of $H_{a}$ is high, the anchor sample is considered as the hard-classified sample (the probability of classifying the sample using the softmax classifier is around $1/N$ for each class). In this case, the number of neighbors $K_a$ should be increased and vice versa. Therefore, ANN effectively integrates both global and local information of the training data into metric embedding learning, which can successfully overcome the problems of slow convergence and poor optima in the triplet loss. Besides, different from the triplet loss (which only exploits the pairwise distance), the ANN loss considers the average distance between the anchor and the neighborhood of positive/negative samples, which makes the training process quickly converge.

	Finally, to learn both separable and discriminative features, we combine the softmax loss (denoted as $ L_{s}$) with the ANN loss to jointly optimize the deep neural network, that is,
\begin{equation}\label{eq:loss}
L = L_{s} + \lambda L_{ANN},
\end{equation}
where $\lambda$ is the tradeoff parameter used to balance the two loss functions. %We empirically set $\lambda$ to be 1 in this paper.

\section{Experiments}
In this section, several person ReID datasets used for evaluation are introduced in Section 3.1. Then, the influence of the parameters is given in Section 3.2. Next, the ablation study is shown in Section 3.3. Finally, the comparison with several state-of-the-art methods is presented in Section 3.4.
\subsection{Datasets}
	To verify the effectiveness of the proposed method, we perform extensive experiments on four challenging person ReID datasets, including Market1501 \cite{Zheng2015Scalable}, DukeMTMC-reID \cite{Ristani2016Performance}, CUHK03 \cite{Li2014DeepReID}, and Partial REID \cite{Zheng2016Partial}. The Market1501 dataset contains 1,501 identities captured by six camera views, where the dataset is split into 12,936 training images with 750 identities and 19,732 gallery images with 750 identities. The DukeMTMC-reID dataset contains 1,404 identities collected  from eight cameras. The dataset is divided into 16,522 training images with 702 identities and 17,661 gallery images with 702 identities. The CUHK03 dataset contains 13,164 images with 1,360 identities
\begin{figure}[!thb]	
\centering
\begin{minipage}[c]{0.5\textwidth}
 \begin{subfigure}{0.5\textwidth}
  \centering
  \includegraphics[width=4.2 cm,height=3.2cm]{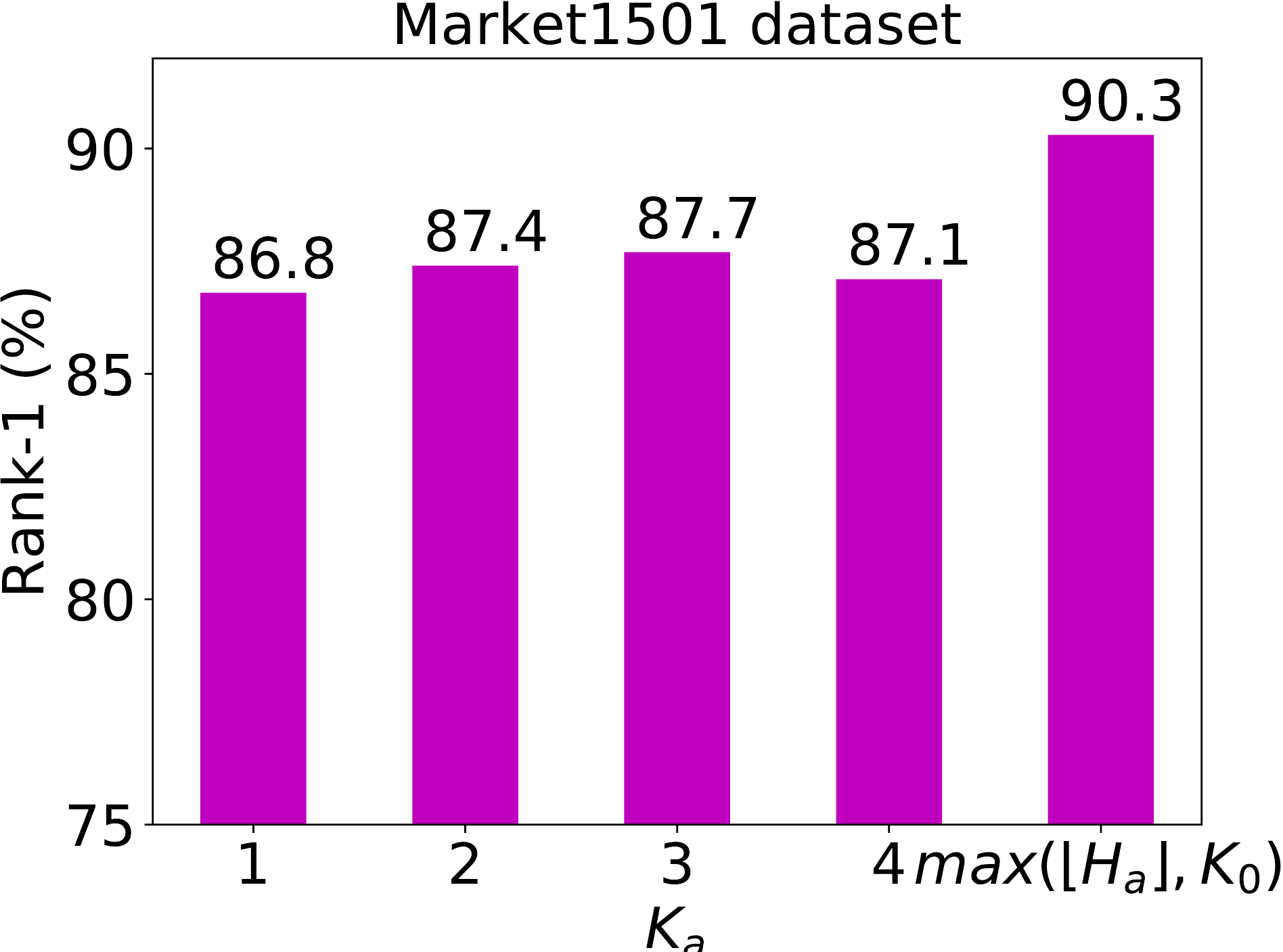}
  \caption{Performance w.r.t. $K_a$}
  \label{fig:k}
\end{subfigure}%
\begin{subfigure}{0.5\textwidth}
  \centering
  \includegraphics[width=4 cm,height=3.2cm]{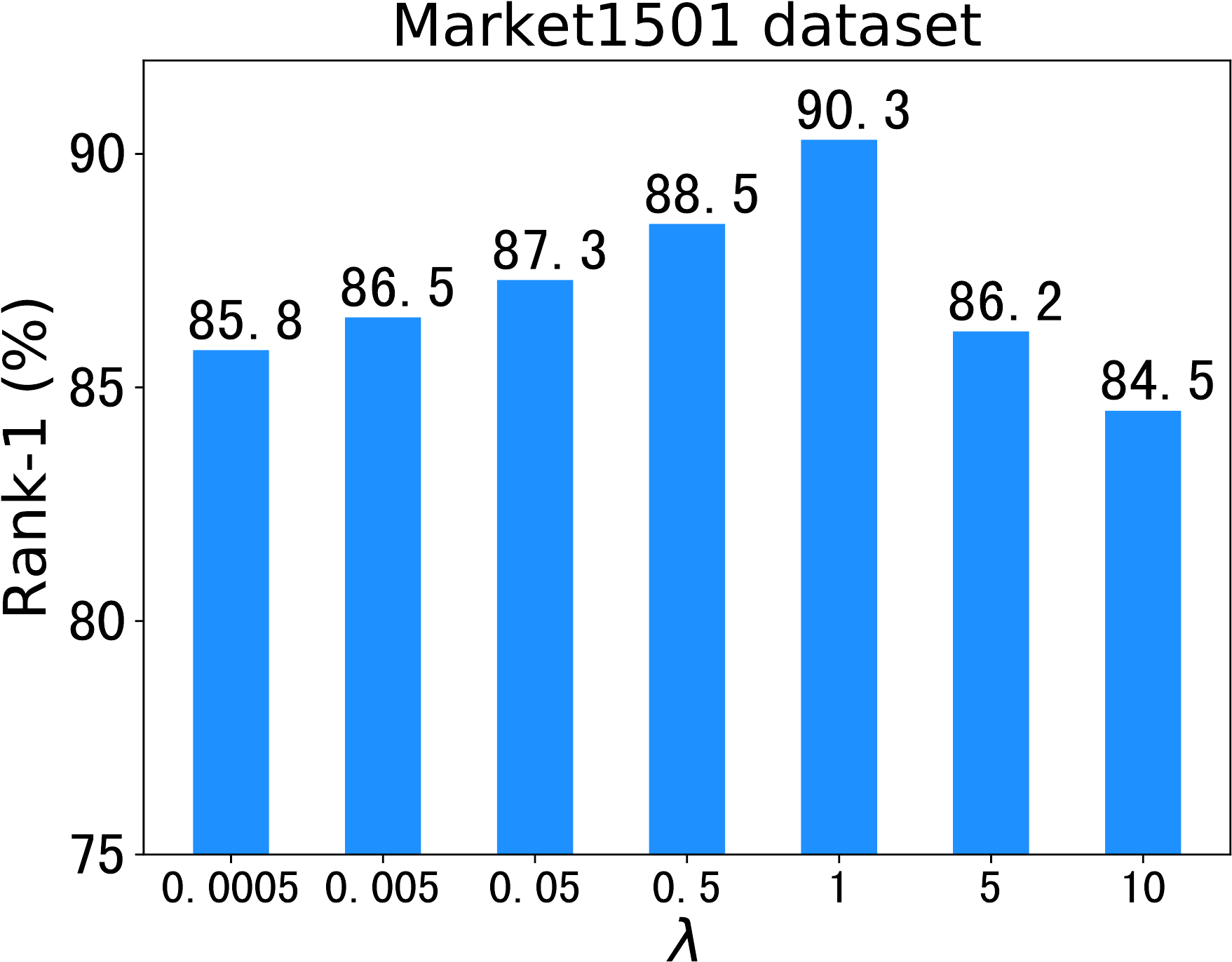}
  \caption{Performance w.r.t. $\lambda$}
  \label{fig:lam}
\end{subfigure}
\caption{The rank-1 (\%) accuracy with different values of (a) $K_a$ and (b) $\lambda$ on the Market1501 dataset.}
\label{fig:hyper}
\end{minipage}
\end{figure}
captured by six cameras. Each identity is observed by two disjoint camera views, yielding an average of 4.8 images in each view. The Patial REID dataset contains 600 images of 60 identities, with 5 full-body images and 5 partial occluded images for each identity.
%The bounding boxes used in the Market1501 and CUHK03 datasets are generated by the deformable part models (DPM) detector \cite{felzenszwalb2008discriminatively}, while they are manually annotated in the DukeMTMC-reID.
%% figure 3

	We use the standard metrics, including the mean Average Precision (mAP) and the Cumulative Matching Characteristic (CMC) curve at rank-1, to evaluate the performance of person ReID.
%\subsection{Implementation Details}
%	We deploy the ResNet \cite{He2016Deep} model pre-trained on ImageNet \cite{Deng2009ImageNet} as the base CNN model to extract mid-level features, where all the convolutional layers are kept and fully-connected layers are removed. We add a global average pooling layer to extract global features (a 512-d vector) and an average pooling layer by each row to extract local features (a 128-d vector). The hidden units of the LSTM layer is set to 128 and the number of LSTM layer is set to 1.
%	We add a fully-connected layer with 128 units for the fused features, following by batch normalization, ReLU, and Dropout. The dropout probability is set to 0.5. For loss function, we empirically set $m$ to 0.3 and $\lambda$ to 0.05. In the train stage, the mini-batch size is set to 128 in which each identity has 4 randomly selected images. In the test stage, we extract the fully-connected layer with 128 units as features for all datasets. All the input images are resized to $256\times128$.
%\begin{figure}[b]
%\centering
%\includegraphics[scale=0.45]{occ_com_market1501}
%\caption{Comparison of the rank-1(\%) accuracy obtained by different variants under different occluded ratios in query images on the Market1501 dataset.}
%\label{fig:occ_com}
%\end{figure}
\subsection{Influence of the parameters}
To observe the influence of the parameters on the proposed method, we evaluate two critical parameters in the proposed method, i.e., the neighbor size (i.e., $K_a$ in Eq.~(\ref{eq:sizeK})) and the tradeoff parameter to combine two losses (i,e., $\lambda$ in Eq.~(\ref{eq:loss})). The rank-1 accuracy with different values of $K_a$ and $\lambda$  on the Market1501 dataset is given in Fig.~2.

From Fig.~2, we can see that the proposed method with the adaptive value of $K_a$ can achieve much better results than that with the fixed value of $K_a$, which demonstrates the importance of adaptive neighborhood.
The value of $\lambda$ also significantly affects the final performance. In summary,
when $\lambda$ is set to 1 and $K_a$ is adaptively set based on the classification uncertainty $H_a$, the proposed method achieves the best performance. In the following, the value of $\lambda$ is fixed to 1.

\subsection{Ablation study}%Roubustness to the Occlusions}
	In this section, we evaluate several variants of the proposed method to verify the effectiveness of the key components in the proposed method for person ReID under occlusions. We conduct the experiments on the Market1501 dataset, where the query images are set with the different levels of occlusions. More specifically, we randomly occlude a region with random values in an image with the aspect ratio $s$. The aspect ratio of occluded area is set within the range of [0.0, 0.6] (please refer to \cite{Zhong2017Random}  for more details). Moreover, we also conduct the experiments on the real and challenging occlusion dataset, the Partial REID dataset, which contains  different types of severe occlusions. The partial occluded images are used as the query images and the full-body images are used as the gallery images.
	
	The proposed method contains two key components: LSTM, which exploits the spatial dependencies between the different local regions to enable the model to be robust to occlusions; and the ANN loss, which learns discriminative metric embeddings.
\begin{table}[!t]
\captionsetup{justification=centering}
\caption{{The details of the nine variants.}}
\label{table:variants}
\centering
\setlength{\tabcolsep}{0.7mm}
\scalebox{0.8}
{
\begin{tabular}{l|l|l|l}
\hline
Variants & Global & Local & Loss \\ \hline
$RN_S$ & \multirow{2}{*}{GAP} & - & Softmax \\ \cline{1-1} \cline{3-4}
$RN_A$ &  & - & Softmax+ANN \\ \hline
\hline
$RNCONV_A$ & \multirow{3}{*}{GAP} & \begin{tabular}[c]{@{}l@{}}Conv\\ ReLU\\ Batch Normalization\end{tabular} & \multirow{3}{*}{Softmax+ANN} \\ \cline{1-1} \cline{3-3}
$RNFC_A$ &  & \begin{tabular}[c]{@{}l@{}}FC\\ ReLU\end{tabular}  &  \\ \cline{1-1} \cline{3-3}
$RNRNN_A$ &  & RNN &  \\ \hline
\hline
$RNLSTM_S$ & \multirow{3}{*}{GAP} & \multirow{3}{*}{LSTM} & Softmax \\ \cline{1-1} \cline{4-4}
$RNLSTM_C$ &  &  & Softmax+Contrastive \\ \cline{1-1} \cline{4-4}
$RNLSTM_T$ &  &  & Softmax+Triplet \\ \hline
\hline
$RNLSTM_A$ (ours) & GAP & LSTM & Softmax+ANN \\ \hline
\end{tabular}
}
\end{table}
	Therefore, nine different variants of the proposed method are evaluated. That is, (1) The baseline method (denoted as $RN_S$) that only uses the global branch and a ResNet model [22] based on the softmax loss. (2) The method (denoted as $RN_A$) that uses the same network as $RN_S$, where both the softmax loss and the ANN loss are employed to jointly optimize the network.
	(3)-(6) The methods (respectively denoted as $RNCONV_A$, $RNFC_A$, $RNRNN_A$, and $RNLSTM_A$) that employ a ResNet model, and combine the global branch and the local branch (here the convolutional layer, fully-connected layer, RNN layer and LSTM layer are respectively used as the local branch), where the softmax loss and the ANN loss are used. Note that $RNLSTM_A$ is the proposed method in this paper.
	(7) The method (denoted as $RNLSTM_S$) that uses the proposed network, where only the softmax loss is used.
	(8)-(9) The methods (respectively denoted as $RNLSTM_C$, $RNLSTM_T$) that uses the proposed network, where the softmax loss is combined with the contrastive loss \cite{Chen2014Deep} and triplet loss \cite{Hermans2017In}, respectively. The details of the nine variants are summarized in Table \ref{table:variants}.
	% The variants $RN_A$ and $RN_S$ are used to evaluated the effectiveness of ANN when only using the global branch. The variants $RNLSTM_S$, $RNLSTM_P$, $RNLSTM_T$ and $RNLSTM_A$ (the proposed method) are used to test the effectiveness of ANN on the proposed framework (using both the global and local branch). The variants $RNCONV_A$,  $RNFC_A$, $RNRNN_A$, and $RNLS TM_A$ (the proposed method) are used to evaluated the importance of using LSTM as the local branch.

%%% table
\begin{table*}[!ht]
\captionsetup{justification=centering}
\caption{{The rank-1 (\%) accuracy and mAP (\%) obtained by the proposed method and the state-of-the-art methods against the different levels of occlusions on the Market1501, DukeMTMC-reID, and CUHK03 (detected) datasets. The best and second highest results are in red and blue, respectively.}}
\label{table:comp}
\centering
\setlength{\tabcolsep}{0.6mm}
\scalebox{0.85}{
{\begin{tabular}{c|cc|cc|cc|cc|cc|cc|cc|cc|cc}
\hline
\multirow{3}{*}{Method} & \multicolumn{6}{c|}{Market1501}                                                    & \multicolumn{6}{c|}{DukeMTMC-reID}                                                 & \multicolumn{6}{c}{CUHK03}                                                        \\ \cline{2-19}
                        & \multicolumn{2}{c|}{s=0} & \multicolumn{2}{c|}{s=0.3} & \multicolumn{2}{c|}{s=0.6} & \multicolumn{2}{c|}{s=0} & \multicolumn{2}{c|}{s=0.3} & \multicolumn{2}{c|}{s=0.6} & \multicolumn{2}{c|}{s=0} & \multicolumn{2}{c|}{s=0.3} & \multicolumn{2}{c}{s=0.6} \\ \cline{2-19}
                             & rank-1         & mAP        & rank-1          & mAP         & rank-1          & mAP         & rank-1  & mAP  & rank-1   & mAP  & rank-1   & mAP  & rank-1  & mAP  &rank-1   & mAP  &rank-1   & mAP  \\ \hline
XQDA \cite{Liao2015Person}   & 43.0        & 21.7       & 28.3         & 14.2        & 24.3         & 12.0        & 31.2 & 17.2 & 20.5  & 10.6 & 17.4  & 9.4  & 44.2 & -    & 36.9  & -    & 32.3  & -    \\
NPD \cite{Zhang2016Learning} & 55.4        & 30.0       & 39.6         & 19.1        & 32.5         & 16.1        & 46.7 & 27.3 & 33.7  & 17.7 & 29.7  & 15.7 & 53.7 & -    & 39.5  & -    & 33.8  & -    \\ \hline
          \hline
IDE \cite{zheng2016person}   & 81.9        & 61.0       & 62.4         & 48.2        & 45.6         & 36.4        & 66.3 & 45.2 & 57.9  & 41.6 & 41.3  & 30.3 & 68.2 & 62.7 & 65.1  & 59.8 & 46.2  & 43.6 \\
TriNet \cite{Hermans2017In}   & 83.2        & 64.9       & \color{blue}68.6         & \color{blue}54.7        & 47.9         & 38.9        & 71.4 & 51.6 & 56.0  & 40.8 & 39.0  & 28.4 & 79.1 & 76.4 & 68.0  & 66.9 & 48.1  & 49.2 \\
PAN \cite{Zhao2017Deeply}    & 81.0        & 63.4       &    52.0 & 36.5            &    43.2     & 30.0     & 71.6 & 51.5 &  44.7   & 29.0   &  39.9   & 25.9      & \color{blue}85.4 & \color{red}90.9 &  61.0  & 66.5  & 53.0  & 57.6      \\
SVDNet \cite{Sun2017SVDNet}  & 81.4        & 61.2       & 62.3         & 46.9        & \color{blue}52.0   & \color{blue}40.3        & 75.9 & 56.3 & \color{blue}59.1  & \color{blue}43.5 & \color{blue}50.6  & \color{blue}37.9 & 81.2 & \color{blue}84.5 &  \color{blue}71.2 &  \color{blue}66.8 & 	\color{red}63.9  & \color{red}62.1      \\
DPFL \cite{chen2017person}   & \color{blue}88.6    & \color{blue}72.6       & -     &  -     &    -   &   -    & \color{red}79.2 &\color{blue} 60.6 &  -    & -    & -      & -     & 82.0 & 78.1 &   -    &  -    &   -  &  -    \\ \hline \hline
$RNLSTM_{A}$      &\color{red} 90.3  & \color{red}76.4    & \color{red}76.6    & \color{red} 63.8   & \color{red}52.9   & \color{red}44.8        & \color{blue} 77.0 & \color{red}62.1 &\color{red} 69.3  & \color{red}58.3 & \color{red}51.9  & \color{red}41.2 & \color{red}86.1 & 83.6 & \color{red}77.0  & \color{red}75.3 & \color{blue}59.8  & \color{blue}59.6 \\ \hline
\end{tabular}
 }
}
\end{table*}

\begin{figure}[!b]
\centering
\begin{minipage}[c]{0.5\textwidth}
 \begin{subfigure}{0.5\textwidth}
  \centering
  \includegraphics[width=4 cm,height=3.2cm]{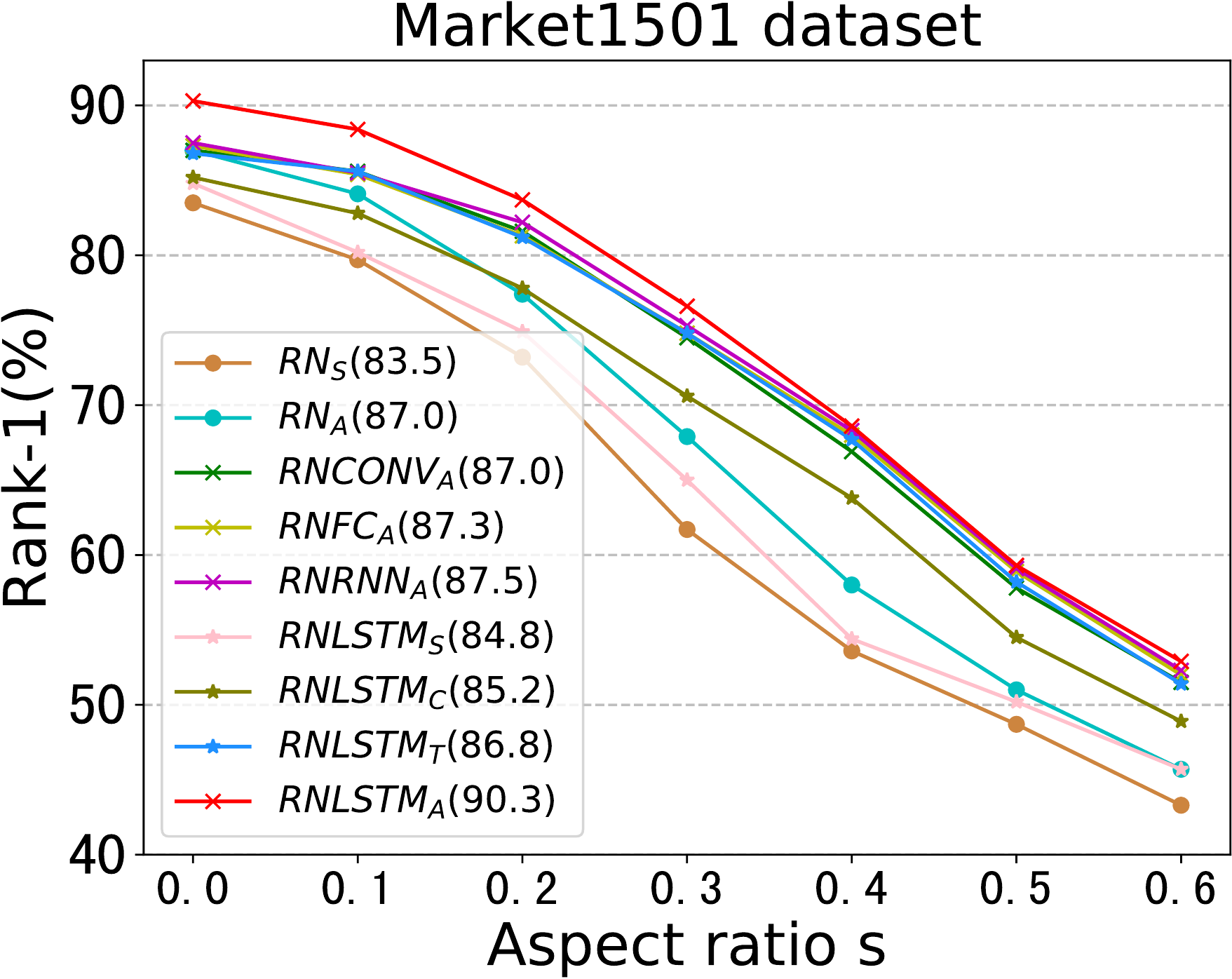}
  \caption{Performance on Market1501.}
  \label{fig:occ_market}
\end{subfigure}%
\begin{subfigure}{0.5\textwidth}
  \centering
  \includegraphics[width=4.2 cm,height=3.2cm]{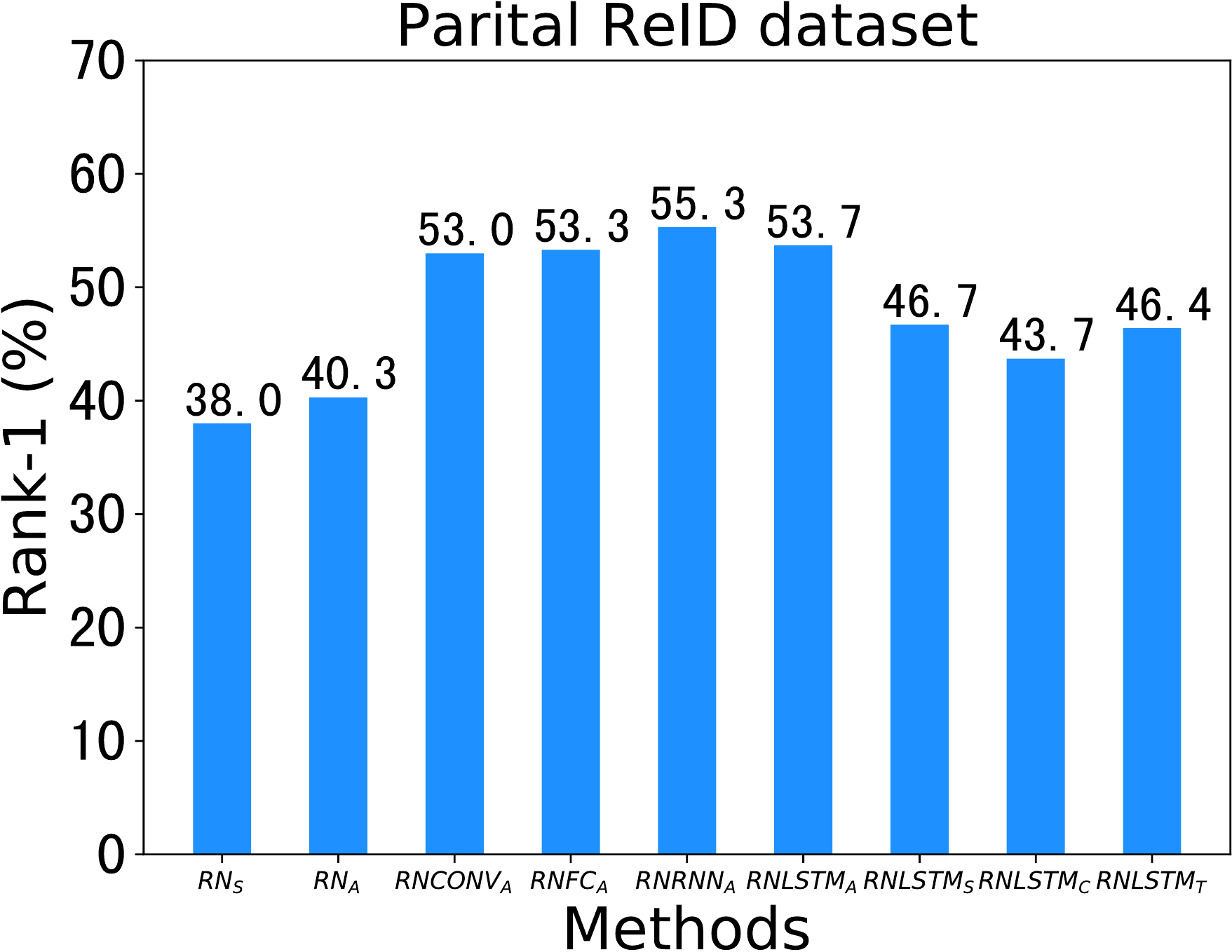}
  \caption{Performance on Partial REID. }
  \label{fig:occ_partial}
\end{subfigure}
\caption{Comparison of the rank-1 (\%) accuracy obtained by different variants on the (a) Market1501 and (b) Partial REID datasets.}
\label{fig:occ_com}
\end{minipage}
\end{figure}

	The rank-1 accuracy obtained by the nine different variants on the Market1501 and Partial REID datasets is shown in Fig.~\ref{fig:occ_com}, where Fig.~3(a) shows the robustness of different variants against the different levels of occlusions.
 %% figure 2

	From  Fig. 3(a), we have the following conclusions: (1) In general, the recognition performance obtained by all the variants on Market1501 drops when the aspect ratio of occluded area increases. This further demonstrates the challenging task of person ReID under occlusions.
	 (2) By comparing the rank-1 accuracy obtained by $RN_S$ and $RN_A$, the ResNet (only with global branch) jointly optimized with the softmax and ANN losses obtains better performance than that only using the softmax loss. This is mainly because that the joint loss enhances the discrimination ability of the model. However, the improvements are not significant, due to the fact that ResNet only considers the high-level semantic information and ignores the local information, which is critical for classification under occlusions.
	 (3) By comparing the variants ($RNCONV_A$, $RNFC_A$, $RNRNN_A$ and $RNLSTM_A$) with different local branches, we can see that LSTM plays a critical role in the final performance. Specifically, compared with the variants without LSTM ($RNCONV_A$, $RNFC_A$, $RNRNN_A$), the variant with LSTM (the proposed $RNLSTM_A$) can effectively improve the rank-1 accuracy about 3\% under different levels of occlusions. This is due to the fact that that LSTM  not only memorizes the spatial correlations between the different local regions, but also reduces the noises caused by partial occlusions.
	(4) The proposed method ($RNLSTM_A$) outperforms the softmax loss based method ($RNLSM_{S}$), and the joint loss using contrastive loss ($RNLSTM_C$) and triplet loss ($RNLSTM_T$) based methods in a reasonable margin (about 3\%\~{}6\% on Market1501). From this comparison, we can see that the combination of the classification loss and metric loss (i.e., contrastive loss or triplet loss) can improve the performance of the model. Furthermore, $RNLSTM_A$ obtains the superiority performance by exploiting the neighborhood information to enlarge the inter-class dispersion while increasing the intra-class compactness.
	 Among all the competing variants, $RNLSTM_A$ consistently achieves the best results on the Market1501 dataset. This indicates that the deep model (based on LSTM and CNN) jointly optimized by the softmax loss and the ANN loss, can effectively enhance the robustness to occlusions.

	From  Fig. 3(b), we can observe similar conclusions on the Partial REID dataset. Note that $RNLSTM_A$ performs slightly inferior than $RNRNN_A$. This is mainly because that the LSTM layer have more parameters than the RNN layer. In other words, to ensure the effectiveness of the proposed method, a large number of training set are preferred to learn the parameters. However, the training set of Partial REID dataset is small (only 300 images are used for training).

\subsection{Comparison with the state-of-the-art methods}
   In this section, we compare the proposed method (i.e., $RNLSTM_A$) with several representative methods, including the traditional metric learning methods (NPD \cite{Zhang2016Learning}, XQDA\cite{Liao2015Person}) and the recently-proposed deep learning methods (IDE \cite{zheng2016person}, TriNet \cite{Hermans2017In}, PAN \cite{Zhao2017Deeply}, SVDNet \cite{Sun2017SVDNet} and DFFL \cite{chen2017person}).

   The rank-1 accuracy and mAP obtained by all the competing methods are shown in Table \ref{table:comp}.
   Compared with the traditional person ReID methods (NPD and XQDA), the deep learning methods achieve significant performance improvements, which show the superiority of deep learning.
   The proposed method obtains much better results than the softmax-loss based IDE \cite{zheng2016person} method and the triplet loss based TriNet \cite{Hermans2017In} method, which demonstrates the effectiveness of the proposed ANN loss.
   Moreover, the proposed method outperforms the part-based method PAN \cite{Zhao2017Deeply} under occlusions, since we exploit the spatial dependencies based on LSTM to learn discriminative representations.
   Compared with SVDNet \cite{Sun2017SVDNet}, the proposed method achieves higher rank-1 accuracy and mAP under small occlusions. However, the proposed method obtains slightly inferior results under large occlusions ($s=0.6$) on the CUHK03 dataset. Although SVD in the FC layer of SVDNet can effectively extract the discriminative information for person ReID, the training complexity of SVDNet is high.
   DPFL [27] that trains a multi-channel network for multi-scale images achieves slightly better performance than the proposed method that only exploits the single scale image on DUKEMTMC-reID. However, the proposed method obtains better results than DPFL on the challenging CUHK03 database, where each pedestrian contains a relatively small number of training images.

\section{Conclusion}
	In this paper, we propose to exploit spatial dependencies based on LSTM to handle the problem of occlusions for person ReID. To better explore the discriminative capability of deep metric embedding, we propose an adaptive nearest neighbor loss to enlarge the inter-class dispersion while preserving the intra-class compactness. Experimental results on four challenging datasets have shown the effectiveness of the proposed method for person ReID under occlusions.
\section*{Acknowledgements}
	This work was supported by the National Key R\&D Program of China under Grant 2017YFB1302400, by the National Natural Science Foundation of China under Grants 61571379, 61503315, U1605252, and 61472334, by the Natural Science Foundation of Fujian Province of China under Grant 2017J01127 and 2018J01576, by the Fundamental Research Funds for the Central Universities under Grant 20720170045, and by State Key Laboratory of Advanced Optical Communication Systems Networks, China.
\section*{References}
\bibliographystyle{elsarticle-num}
\bibliography{refs/paper}

 \end{document}